 \documentclass[tablecaption=bottom,pmlr]{jmlr}

\usepackage{booktabs}
\usepackage[load-configurations=version-1]{siunitx} 

\newcommand{\ie}{\textit{i}.\textit{e}.}

\usepackage{amsmath}
\usepackage{dsfont}

\usepackage{paralist}
\usepackage{multirow}
\usepackage{graphicx}
\usepackage[ruled]{algorithm2e}
\SetKwComment{Comment}{$\triangleright$\ }{}
\usepackage{tikz}
\usepackage{enumitem}
\usetikzlibrary{positioning}
\usepackage{wrapfig,booktabs}

\theorembodyfont{\upshape}
\theoremheaderfont{\scshape}
\theorempostheader{:}
\theoremsep{\newline}

\jmlrvolume{148}
\jmlryear{2021}
\jmlrsubmitted{submission date}
\jmlrpublished{publication date}
\jmlrworkshop{NeurIPS 2020 Preregistration Workshop} 

\title{Robustness May Be at Odds with Fairness: An Empirical Study on Class-wise Accuracy}

  \author{\Name{Philipp Benz\nametag{\thanks{Equal contribution}}} \Email{pbenz@kaist.ac.kr} \\
   \Name{Chaoning Zhang}$^{*}$ \Email{chaoningzhang1990@gmail.com}\\
   \Name{Adil Karjauv} \Email{mikolez@gmail.com} \\
   \Name{In So Kweon} \Email{iskweon77@kaist.ac.kr} \\
   \addr Korea Advanced Institute of Science and Technology (KAIST)}

\begin{document}

\maketitle

\begin{abstract}
Convolutional neural networks (CNNs) have made significant advancement, however, they are widely known to be vulnerable to adversarial attacks. Adversarial training is the most widely used technique for improving adversarial robustness to strong white-box attacks. Prior works have been evaluating and improving the model average robustness without class-wise evaluation. The average evaluation alone might provide a false sense of robustness. For example, the attacker can focus on attacking the vulnerable class, which can be dangerous, especially, when the vulnerable class is a critical one, such as ``human'' in autonomous driving. We propose an empirical study on the class-wise accuracy and robustness of adversarially trained models. We find that there exists inter-class discrepancy for accuracy and robustness even when the training dataset has an equal number of samples for each class. For example, in CIFAR10, ``cat'' is much more vulnerable than other classes. Moreover, this inter-class discrepancy also exists for normally trained models, while adversarial training tends to further increase the discrepancy. Our work aims to investigate the following questions: (a) is the phenomenon of inter-class discrepancy universal regardless of datasets, model architectures and optimization hyper-parameters? (b) If so, what can be possible explanations for the inter-class discrepancy? (c) Can the techniques proposed in the long tail classification be readily extended to adversarial training for addressing the inter-class discrepancy? 
\end{abstract}


\section{Introduction}
Convolutional neural networks (CNNs)~\citep{lecun2015deep} have achieved enormous success in a wide range of vision applications~\citep{zhang2019revisiting,kim2019recurrent,kim2020video,pan2020unsupervised,zhang2020deepptz,zhang2020resnet,kim2020Devil,zhang2020udh}. However, they are still widely known to be vulnerable to adversarial examples~\citep{szegedy2013intriguing,goodfellow2014explaining}. Numerous endeavors have been attempted to improve adversarial robustness of deep classifiers, and adversarial training, to our best knowledge, is the only one that has not been broken by strong white-box attacks~\citep{goodfellow2014explaining,madry2017towards,carlini2017towards}. Prior works mainly report the model accuracy and robustness averaging on samples from all classes without class-wise evaluation. This average performance alone might be misleading for giving a wrong sense of robustness. For example, in autonomous driving, a well-performing model with high accuracy and/or robustness averaging on all classes is particularly dangerous if a certain important class, such as \emph{human}, is vulnerable. 

\begin{figure}[t]
\centering
    \includegraphics[width=0.4\linewidth]{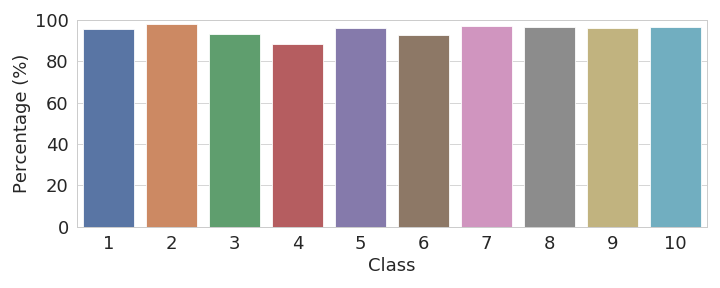}
    \includegraphics[width=0.4\linewidth]{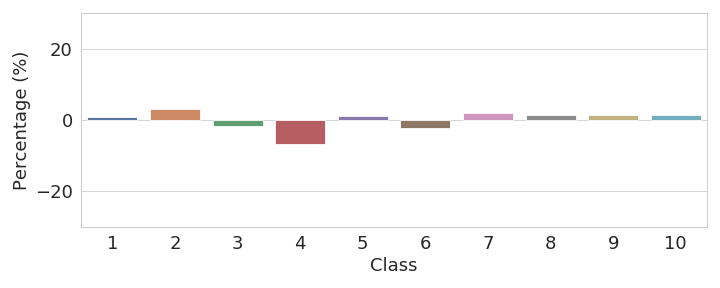}
    \includegraphics[width=0.4\linewidth]{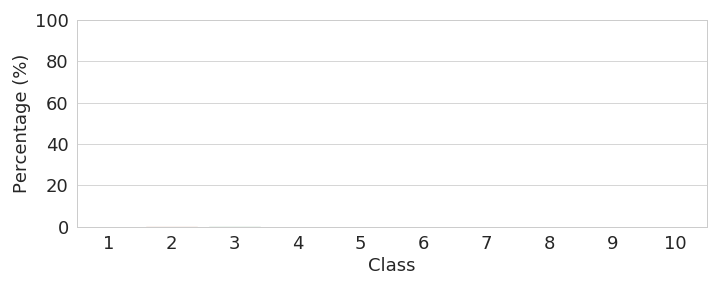}
    \includegraphics[width=0.4\linewidth]{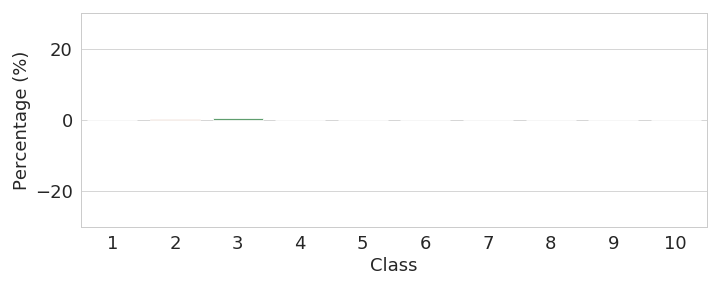}
    \caption{Inter-class discrepancy for standard model. First row: accuracy w/o (left) and w/ (right) mean subtracted. Second row: robustness w/o (left) and w/ (right) mean subtracted.}
    \label{fig:main_standard}
\end{figure}

\begin{figure}[t]
\centering
    \includegraphics[width=0.4\linewidth]{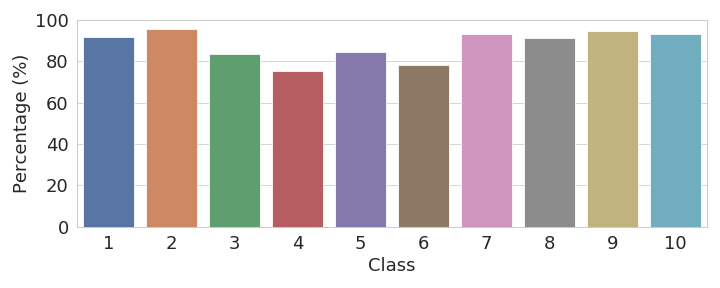}
    \includegraphics[width=0.4\linewidth]{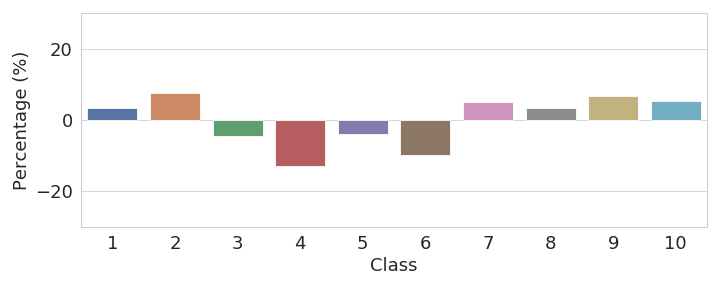}
    \includegraphics[width=0.4\linewidth]{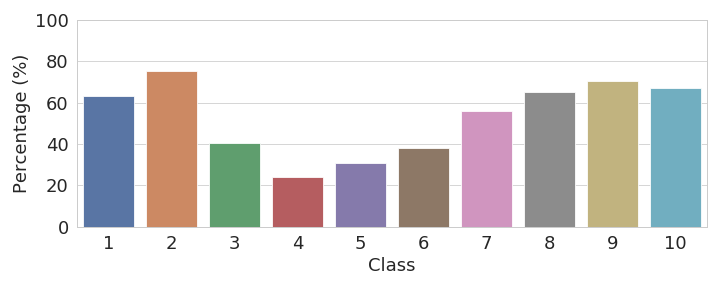}
    \includegraphics[width=0.4\linewidth]{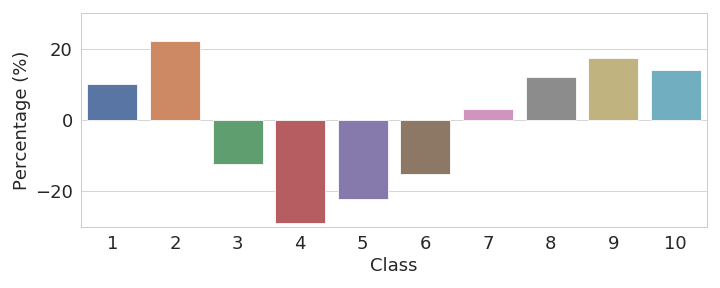}
    \caption{Inter-class discrepancy for adversarially trained model. First row: accuracy w/o (left) and w/ (right) mean subtracted. Second row: robustness w/o (left) and w/ (right) mean subtracted.}
    \label{fig:main_adv}
\end{figure}

Recognizing its practical relevance, we perform an empirical study to evaluate the class-wise accuracy and robustness of adversarially trained models. Preliminary investigation is conducted for CIFAR10 on ResNet18~\citep{he2016deep} with both standard training (see Figure~\ref{fig:main_standard}) and adversarial training (see Figure~\ref{fig:main_adv}). There are several intriguing observations. First, there is a non-trivial inter-class discrepancy, even though the long-tail issue does not exist, \ie\ each class is balanced with the same number of training samples. Second, a similar trend can be observed for the adversarially trained model, more notably, the inter-class discrepancy is more significant than that of a standard model. Third, the imbalance is the most significant for the adversarially trained model. Overall, it suggests that there exists inter-class discrepancy under balanced training dataset and adversarial training makes the inter-class discrepancy more serious. Beyond the already widely recognized importance on the model robustness, we argue that fairness in terms of decreasing the inter-class discrepancy also deserves the attention from the machine learning community.

Our empirical analysis will address the following questions regarding class-wise accuracy and robustness:

\begin{itemize}[noitemsep, topsep=0pt, leftmargin=*]
    \item Is the phenomenon of inter-class discrepancy universal in other setups?
    \item What are possible explanations for this inter-class discrepancy in accuracy and robustness?
    \item Can the techniques proposed in the long-tail setup be readily extended to adversarial training for addressing the inter-class discrepancy?
\end{itemize}


\section{Related work}
\subsection{Adversarial examples}
CNNs are widely known to be vulnerable to adversarial examples~\citep{szegedy2013intriguing,goodfellow2014explaining,ilyas2019adversarial,benz2020batch,benz2020revisiting,benz2020data}, which has inspired numerous investigations on both image-dependent attack~\citep{szegedy2013intriguing,goodfellow2014explaining,madry2017towards} and universal attack~\citep{moosavi2017universal,zhang2019cd-uap,zhang2020understanding,benz2020double,zhang2021universal,zhang2021survey,benz2021universal} and defense~\citep{qiu2019review,yuan2019adversarial}. Most of the defense techniques have been broken, and currently, adversarial training~\citep{goodfellow2014explaining,madry2017towards} is the most widely adopted one, empirically proven effective. In the past few years, numerous adversarial training methods~\citep{zhang2019you,wang2019bilateral,shafahi2019adversarial,zhang2019theoretically,wong2020fast} have been proposed for improving either effectiveness or efficiency. Despite different motivations and implementations, they all fall into a min-max optimization problem~\citep{madry2017towards}, \ie\ adversarial attack solving the loss maximization problem to generate adversarial examples and network training minimizing the loss to update the network weights. Adversarial training leads to some interesting findings. For example, adversarial training leads to higher robustness while at the cost of accuracy, inspiring to bridge the gap for the trade-off between robustness and accuracy~\citep{zhang2019theoretically}. Adversarial training leads to a model with more robust features~\citep{ilyas2019adversarial}, consequently reducing information while improving transferability~\citep{terzi2020adversarial}. Adversarially trained models have also been found to be more fit for down-stream transfer learning~\citep{utrera2020adversarially}. Complementary to their findings, we empirically find that adversarial training increases inter-class discrepancy for accuracy and robustness.  

\subsection{Inter-class discrepancy in long tail recognition}
The inter-class discrepancy problem in long-tail recognition is a fundamental issue in machine learning. In the real world setting, the long-tail problem exists when data is inherently imbalanced \citep{kendall1946advanced}, which undermines the performance of algorithms that do not take this problem into account \citep{buda2018systematic, japkowicz2002class}. Due to its practical relevance, there is a body of work devoted to tackling this problem \citep{he2009learning}. The core issue in the long tail recognition lies in low accuracy for the rare classes, and various techniques have been developed to improve the accuracy of those rare classes. Our results suggest that there exists an inter-class discrepancy for accuracy and robustness for a balanced dataset, especially, adversarial training leads to a more significant inter-class discrepancy. Conceptually, the vulnerable class in the adversarial training is similar to the rare class in the long tail recognition, since the accuracy of them is low and needs to be improved. Straightforwardly, the techniques proposed to solve the long tail recognition might also help mitigate the inter-class discrepancy for the accuracy and/or robustness of adversarially trained models.
There are two common techniques in long-tail recognition for handling a class-imbalanced dataset. The first one is to modify the dataset to reduce the imbalance. One can either collect more data samples for the deficient classes \citep{chawla2002smote, he2008adasyn} or remove samples from the abundant classes to increase balance \citep{drummond2003c4}. Alternatively, the sampling strategy can be designed to increase the sampling frequency for the rare classes during training. The second technique is called cost-sensitive learning, which modifies misclassification costs to account for the imbalance in the number of samples \citep{tang2008svms, elkan2001foundations, huang2016learning}. A recent method \citep{cui2019class} introduces the weighting factors for each class to re-balance the loss function. These weights are inversely proportional to the effective number of samples for every class. \citep{cao2019learning} proposes a new re-balancing optimization procedure with a new loss function to encourage larger classification margins for deficient classes.

\section{Methodology and experimental protocol}
\subsection{Is the phenomenon of inter-class discrepancy universal in other setups?}
To test how universal the inter-class discrepancy for accuracy and robustness is, we mainly take into account three factors, \ie\ datasets, model architectures and optimization methods. For dataset, we plan to test it on various benchmark datasets, including MNIST \citep{lecun1998gradient}, SVHN \citep{netzer2011reading}, CIFAR10 \citep{krizhevsky2009learning}, CIFAR100 \citep{krizhevsky2009learning}, TinyImageNet, ImageNet \citep{deng2009imagenet}. For models, we plan to evaluate on the most seminal models ranging from networks stacking a few convolutional layers to very deep networks, specifically including LeNet \citep{lecun1998gradient}, VGG family (VGG16 and VGG19) \citep{simonyan2014very}, ResNet family (ResNet18 and ResNet50) \citep{he2016deep}, DenseNet family (DenseNet121 and DenseNet169) \citep{huang2017densely}, Inception family (GoogleNet, Inception-v3) \citep{szegedy2015going}. For optimization factors, we mainly consider optimizer, and learning rate schedule and weight decay. 

To check whether the same phenomenon can be observed on the above datasets, we will adopt ResNet18 to train a standard and adversarially trained model for each dataset and evaluate the class-wise accuracy and robustness. For the ImageNet dataset, to avoid expensive computation, we will use the online available pre-trained ResNet50, for both standard and adversarially trained ones. For various models, we will test them only on CIFAR10 dataset for avoiding redundancy. For the optimization factors, we will test them with CIFAR10 on ResNet18, but with different optimizers, such as SGD and ADAM, different learning rate schedule (stepwise decrease and cyclic), and different weight decay factors.

Additionally, it would be interesting to see whether the vulnerable class changes during the training. Thus, we will also report the trend of class-wise accuracy and robustness during the whole training stage. 

\subsection{What might be possible explanations for the inter-class discrepancy?}
At this stage, we assume that the results from the above investigation would support the following conclusion: The phenomenon of inter-class discrepancy should exist universally for a wide range of datasets on various model architectures with different optimization hyper-parameters.  In the balanced dataset setup, each class has the same number of training samples, which increases the chance that the model architectures and/or optimization strategies might influence this phenomenon. We aim to analyze the following results:

\begin{itemize}[noitemsep, topsep=0pt, leftmargin=*]
    \item Is the same trend of inter-class discrepancy observed for different networks trained on the same dataset, \ie\ a robust/vulnerable class on \emph{model A} is also robust/vulnerable on \emph{model B}? If so, we can conclude that the inter-class discrepancy has little to do with model architectures. Otherwise, model architecture can be one factor that causes the phenomenon of the inter-class discrepancy.
    \item Similarly, we can check the trend regarding different optimization factors. If the same trend is observed for models trained with different optimization, such as standard training vs.\ adversarial training, or SGD vs.\ ADAM, we can conclude that this inter-class discrepancy has little to do with optimization factors. Otherwise, optimization factors can be one factor influencing the inter-class discrepancy.
\end{itemize}

Another important factor is the semantic features in a dataset. We can conduct experiments on CIFAR100, which has super-classes, under which multiple sub-classes share similar semantic features. We can visualize a matrix of ground-truth classes and predicted classes. For example, for samples from ground-truth \emph{class x}, we can count the number of predicted classes and the majority of the samples are likely to be predicted as \emph{x}. If \emph{class x} is semantically similar to \emph{class y}, we expect that there would be a non-trivial amount of samples misclassified to be \emph{class y}. Vice versa, there would also be a nontrivial amount of samples from ground-truth \emph{class y} misclassified to be \emph{class x}. Cross-class feature similarity, represented by the cosine similarity between the output logit vector of two different ground-truth labels, might be a good metric to measure the similarity between classes. Specifically, we measure the average logit vector for samples from each ground-truth label and then perform cross-class feature cosine similarity. If a label has high cross-class feature similarity with other labels, it means the model perceives it to be close to other labels, which might lead to the samples from this label being vulnerable to misclassification. 

\textbf{Feature perspective.} Recently, adversarial robustness has been attributed to non-robust features in the dataset~\citep{ilyas2019adversarial}.
It has been shown in~\citep{ilyas2019adversarial} that both robust features and non-robust features are useful for classification. It would be interesting to distinguish whether robust or/and non-robust features lead to the inter-class discrepancy. Following~\citep{ilyas2019adversarial}, we will construct a dataset that has non-robust features, and train a new model on them and perform class-wise accuracy evaluation. As a control study, we will also construct a robust dataset and repeat the above procedure.

\subsection{Can the techniques from long tail be extended to adversarial training?}
Since our setup has a balanced number of samples for each class, we mainly evaluate the cost-sensitive learning strategy, by giving higher weight on the vulnerable class(es). Since we do not know the performance yet, it is challenging to provide more concrete procedures. We take the following two scenarios into account. First, we assume each class is equally important for the model and the target is to decrease the inter-class discrepancy while minimizing the decrease of the overall average accuracy and/or robustness. Second, we assume that a certain class is critical and the target is to increase the accuracy and/or robustness for that class while minimizing the decrease of accuracy/robustness for other classes. Additionally, we consider including a regularizer term to decrease the inter-class cosine similarity. For adversarial training, we will experiment with a targeted attack by choosing the vulnerable or important class as the target class, which intuitively might make that class more robust. Depending on the performance, we will tailor the above strategies accordingly.  

\subsection{Additional explorations for robustness against natural corruptions}
It has been shown in~\citep{ford2019adversarial} that corruptions such as noise corruptions, fog or contrast influence standard and adversarially trained models differently. It would be insightful to see whether they might influence the class-wise performance differently. Moreover, comparing and analyzing the behavior between adversarial perturbation and natural corruptions can provide insight into the phenomenon of the inter-class discrepancy. 

\section{Experimental Results}
The content above is the same as that in the original proposal except for some fixed minor language issues. The content below presents the experimental results that follow the protocol in the proposal. To improve readability, we structure the experimental results in the same order as the proposal.

\subsection{The Universality of the Class-wise Accuracy Phenomenon}
\begin{figure}[htpb!]
    \centering
    \includegraphics[width=0.3\linewidth]{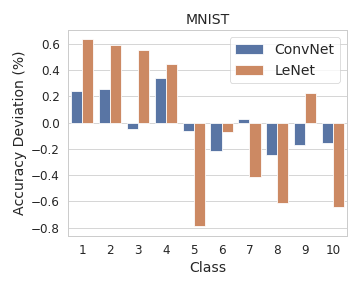}
    \includegraphics[width=0.3\linewidth]{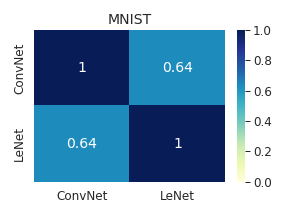}
    \includegraphics[width=0.65\linewidth]{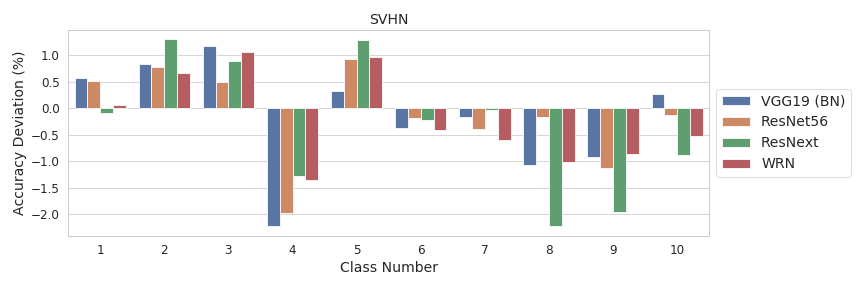}
    \includegraphics[width=0.34\linewidth]{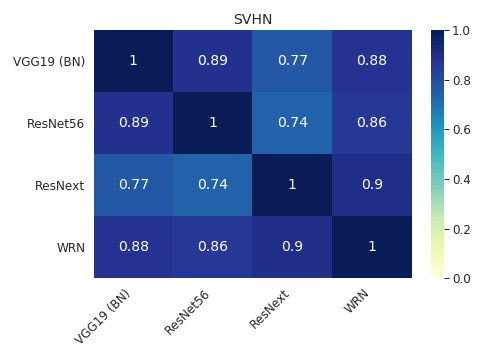}
     \caption{Class-wise deviation from the average accuracy (left) and cosine similarity of the class-wise accuracies of different model architectures (right) on digit classification datasets MNIST (top) and SVHN (bottom).}
\label{fig:clean_model_barplot_numbers}
\end{figure}

\begin{figure}[htpb!]
    \centering
    \includegraphics[width=0.65\linewidth]{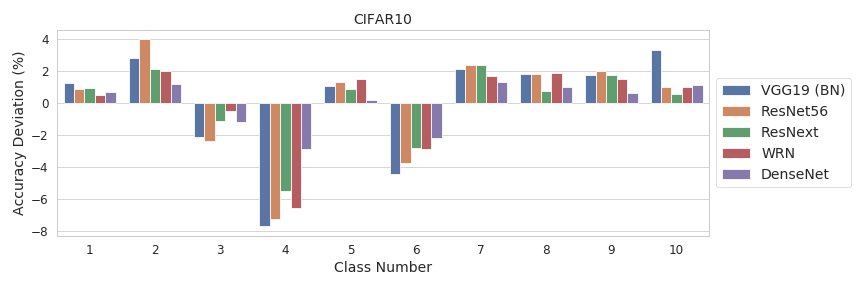}
    \includegraphics[width=0.34\linewidth]{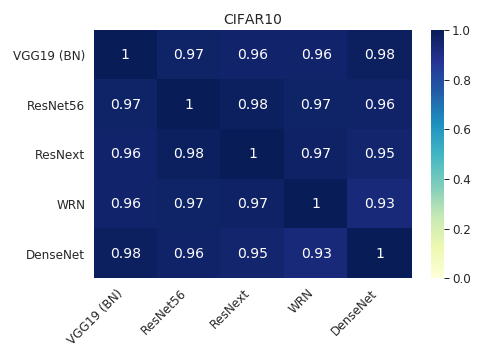}
    \includegraphics[width=0.65\linewidth]{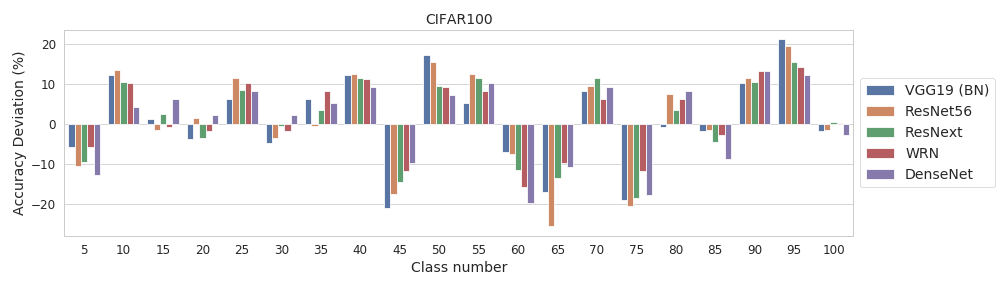}
    \includegraphics[width=0.34\linewidth]{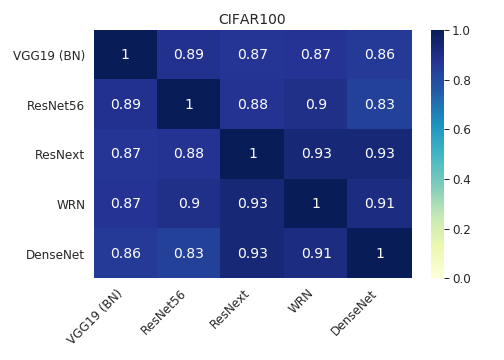}
    \caption{Class-wise deviation from the average accuracy (left) and cosine similarity of the class-wise accuracies of different model architectures (right) for different network architectures trained on CIFAR10 (top) and CIFAR100 (bottom).}
    \label{fig:clean_model_barplot_images_cifar}
\end{figure}

\begin{figure}[htpb!]
    \centering
    \includegraphics[width=0.65\linewidth]{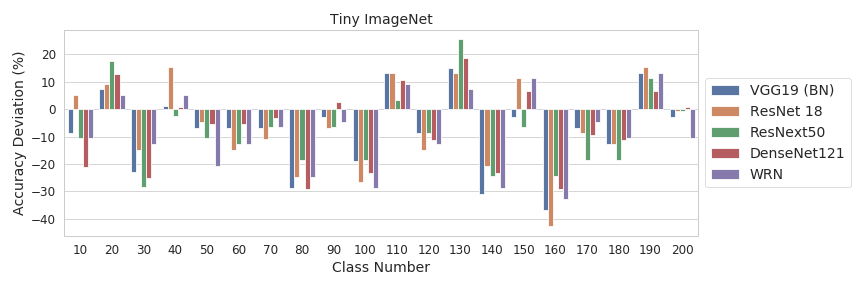}
     \includegraphics[width=0.34\linewidth]{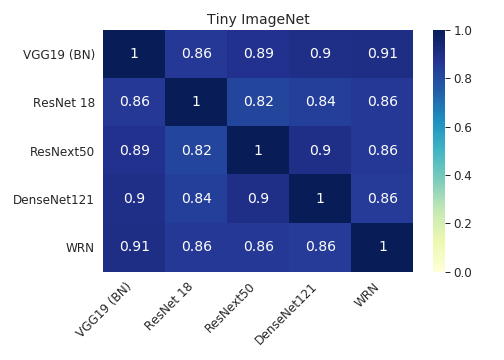}
    \includegraphics[width=0.65\linewidth]{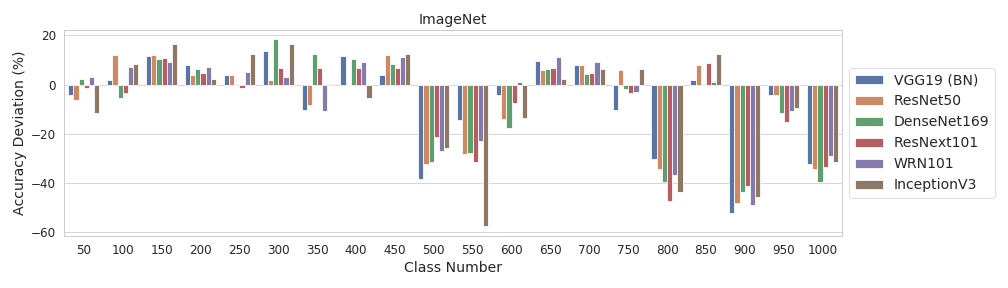}
    \includegraphics[width=0.34\linewidth]{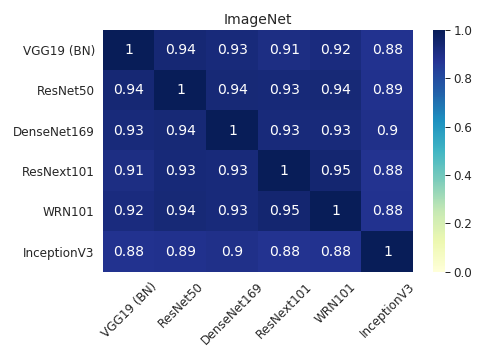}
    \caption{Class-wise deviation from the average accuracy (left) and cosine similarity of the class-wise accuracies of different model architectures (right) for different network architectures trained on Tiny-ImageNet (top) and ImageNet (bottom).}
    \label{fig:clean_model_barplot_images_imagenet}
\end{figure}

\textbf{Dataset and Model Architectures.} 
We report the class-wise phenomenon for different model architectures trained on different datasets. Since we are mainly interested in the inter-class discrepancy, we report the accuracy deviations from the average accuracy for each class as well as the cosine similarities between the distributions of the different models trained on a given dataset. Figure~\ref{fig:clean_model_barplot_numbers} shows the results on the digit classification datasets MNIST and SVHN. For different model architectures, a similar trend regarding the accuracy deviation can be observed. For MNIST, the LeNet architecture shows overall higher deviations than the ConvNet architecture. On SVHN, apart from a few classes, most architectures exhibit similar tendencies. This can be further observed from the cosine similarity matrices. The cosine similarity between the class-wise distribution between ConvNet and LeNet is $0.64$ on MNIST. The average cosine similarity on SVHN is close to 0.84. 
Figure~\ref{fig:clean_model_barplot_images_cifar} and Figure~\ref{fig:clean_model_barplot_images_imagenet} show the results on CIFAR and ImageNet image classification datasets, respectively. For the image classification datasets, it is more apparent that different model architectures show very similar trends for different classes. For example in Figure~\ref{fig:clean_model_barplot_images_cifar} for CIFAR10 and CIFAR100 all cosine similarity values are above 0.93 and 0.83, respectively. In Figure~\ref{fig:clean_model_barplot_images_imagenet} for the ImageNet dataset with 1000 classes the cosine similarity of the class-wise accuracy distributions is always higher than 0.88. Wide-ResNet and ResNext exhibit the highest similarity with a value of 0.95. 

\begin{figure}[htpb!]
    \centering
    \includegraphics[width=0.5\linewidth]{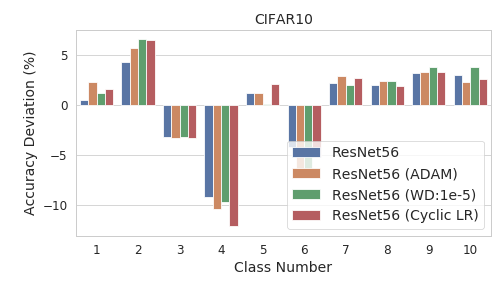}
    \includegraphics[width=0.4\linewidth]{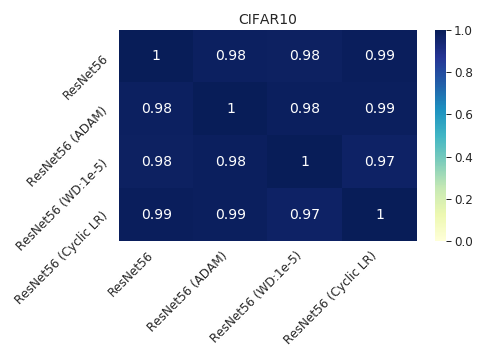}
    \caption{Influence of optimization factors on the class-wise accuracy distribution. The optimizer, weight decay, and learning rate schedule are investigated.}
    \label{fig:optim_influence}
\end{figure}

\textbf{Influence of Optimization Factors.} Here, we investigate the influence of gradient optimizer, learning rate schedule, and weight decay. The baseline was trained with stochastic gradient descent (SGD) and a weight decay of $1e^{-4}$. As an alternative to SGD, we test a model trained with the ADAM~\citep{kingma2014adam} optimizer. Instead of a weight decay of $1e^{-4}$, we train an additional model with a weight decay of $1e^{-5}$. To test the influence of the learning rate schedule, we experiment with the cyclic learning rate~\citep{smith2017cyclical}. The accuracy deviation and cosine similarity plot are shown in Figure~\ref{fig:optim_influence}. It can be observed that none of the investigated optimization techniques has a significant influence on the class-wise accuracies, which is indicated by the cosine similarities being consistently higher than $0.97$. 

\begin{figure}[htpb!]
    \centering
    \includegraphics[width=0.33\linewidth]{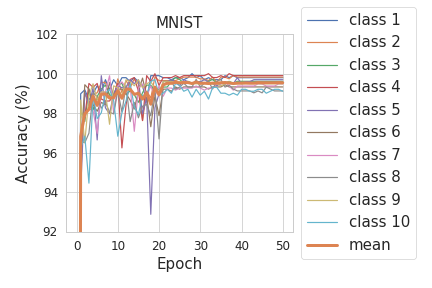}
    \includegraphics[width=0.33\linewidth]{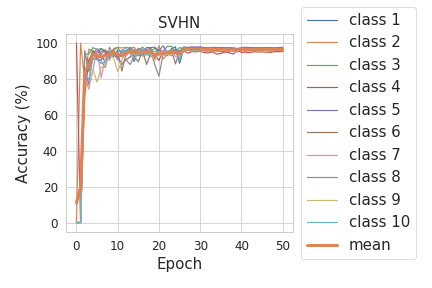}
    \caption{Class-wise accuracies during model training for digit classification.}
    \label{fig:during_training_digit}
\end{figure}

\begin{figure}[htpb!]
    \centering
    \includegraphics[width=0.32\linewidth]{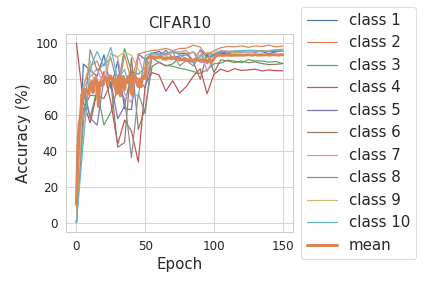}
    \includegraphics[width=0.32\linewidth]{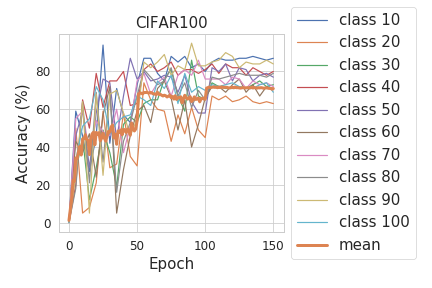}
    \includegraphics[width=0.32\linewidth]{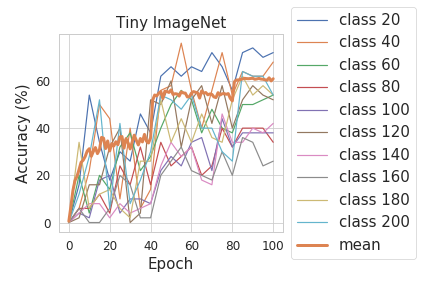}
    \caption{Class-wise accuracies during model training for image classification.}
    \label{fig:during_training_image}
\end{figure}

\textbf{Evolution of Class-wise Accuracy Distribution During Training.}
Figure~\ref{fig:during_training_digit} shows the evolution of the class-wise accuracies during training for the task of digit classification and Figure~\ref{fig:during_training_image} for image classification. It can be observed that in the early stages of training, the class-wise accuracies are very noisy and fluctuate significantly. In the latter stage of training with a smaller learning rate, there still exists the inter-class discrepancy which, however, becomes more stable. For example, for MNIST, class 10 stays being the most vulnerable in the later epochs of training. 

\begin{figure}[htpb!]
    \centering
    \includegraphics[width=0.7\linewidth]{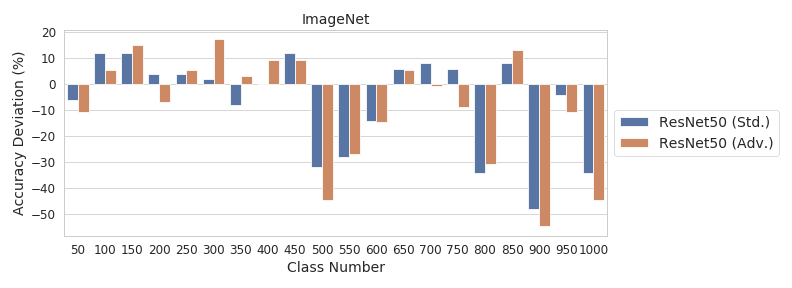}
    \includegraphics[width=0.29\linewidth]{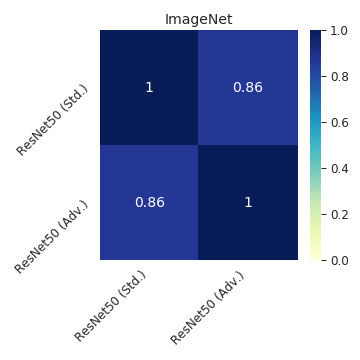}
    \caption{Class-wise deviation from the average accuracy (left) and cosine similarity (right) of a standard trained and adversarially trained ResNet50 architecture trained on ImageNet.}
    \label{fig:comp_adv_std}
\end{figure}

\begin{figure}[htpb!]
    \centering
    \includegraphics[width=1.0\linewidth]{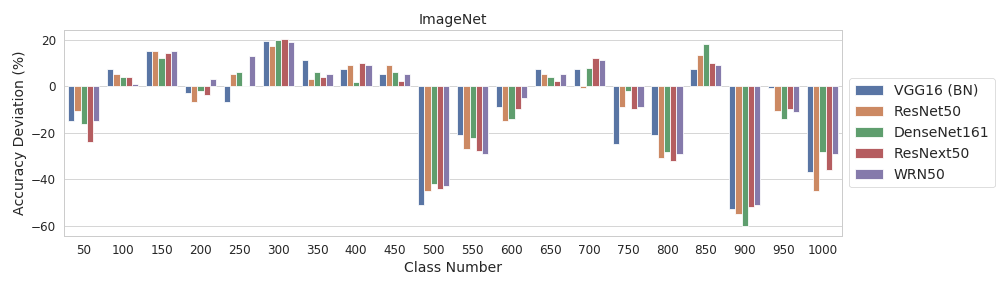}
    \caption{Class-wise deviation from the average accuracy of adversarially trained architectures trained on ImageNet.}
    \label{fig:robust_model_barplot}
\end{figure}

\begin{figure}[htpb!]
    \centering
    \begin{minipage}{.47\textwidth}
        \includegraphics[width=\linewidth]{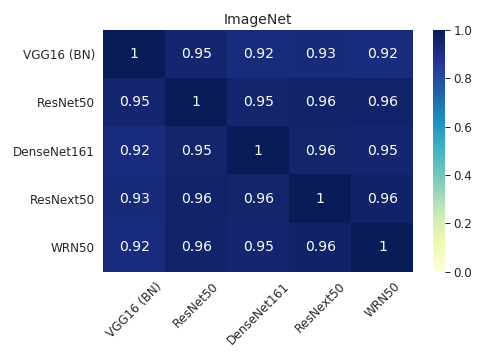}
        \caption{Cosine similarity between various architectures trained on ImageNet.}
    \label{fig:robust_model_heatmap}
    \end{minipage}
    \quad
    \begin{minipage}{.40\textwidth}
        \includegraphics[width=\linewidth]{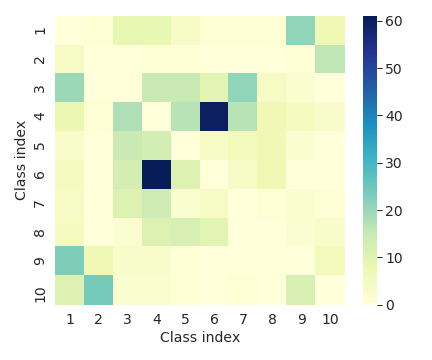}
        \caption{Confusion matrix of CIFAR10 on ResNet56.} 
        \label{fig:cifar10_confusion}
    \end{minipage}
\end{figure}

\textbf{Adversarially Trained Models.} We further investigate the class-wise accuracies and robustness of adversarially trained models. Figure~\ref{fig:comp_adv_std} shows the comparison of a standard trained ResNet50 and adversarially trained ResNet50 as well as their cosine similarities. We observe that even between a normally trained model and an adversarially trained model a similar distribution can be observed. However, adversarially trained models usually exhibit higher class-wise discrepancies than models with standard training, as already indicated in our preliminary results in Figure~\ref{fig:main_standard} and Figure~\ref{fig:main_adv}. The results suggest that robustness might be at odds with class-wise fairness. The class-wise accuracy deviations of adversarially trained models are shown in Figure~\ref{fig:robust_model_barplot} for $5$ different adversarially trained models which are trained with PGD~\citep{madry2017towards} bounded in the $l_2$-ball with an allowed perturbation magnitude of $\epsilon=3.0$~\citep{salman2020adversarially}. From the cosine similarity plot in Figure~\ref{fig:robust_model_heatmap}, it can be observed that all investigated adversarially trained models exhibit a high similarity, above $0.92$, which is overall higher than that observed for the standard trained models in Figure~\ref{fig:clean_model_barplot_images_imagenet}. The results indicate that the similarity between robust models is higher than that of non-robust models, suggesting that robust models seem to learn similar robust features.

\subsection{Possible Explanations of the Class-wise Accuracy Discrepancy}
Our above results demonstrate that the class-wise discrepancy phenomenon is universal in various setups regardless of the dataset classification type, model architectures, optimization factors, etc. Moreover, we find that on the same dataset, the cosine similarity between different models is always very high, indicating the class-wise discrepancy lies in the feature distribution of the dataset instead of model architectures or optimization factors. Thus, we conjecture that the reason for low accuracy or robustness for a certain class can be attributed to the fact that another class having very similar features. For example, in CIFAR10, the class ''cat" and ''dog" have the lowest accuracies because they share more semantically similar features than other classes. To illustrate this, we plot a confusion matrix in Figure~\ref{fig:cifar10_confusion} where class 4 and 6 represent ''cat" and ''dog", respectively. In the shown confusion matrix, the numbers indicate how many samples from the row class index are classified as the column class index. It is worth highlighting that the matrix shows a highly symmetric pattern, demonstrating the bidirectional confusion (or misclassification) between two classes often has a similar probability. 

\begin{figure}[htpb!]
    \centering
    \includegraphics[width=0.3\linewidth]{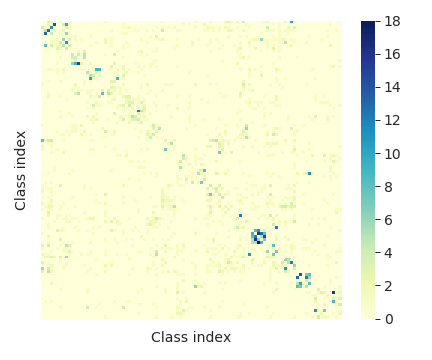}
    \includegraphics[width=0.3\linewidth]{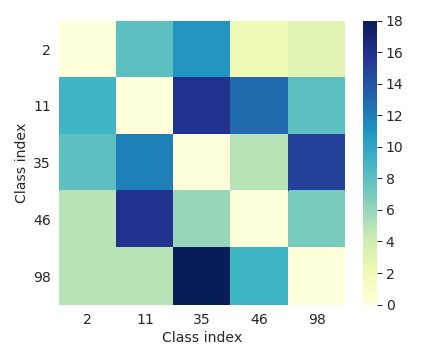}
    \includegraphics[width=0.3\linewidth]{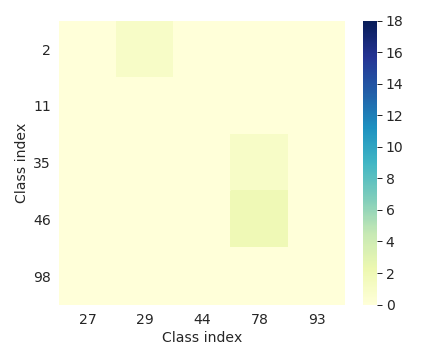}
    \caption{Confusion matrices for a ResNet56 trained on ImageNet for the entire dataset (left), and its super-classes ordered version (right).}
    \label{fig:confusion}
\end{figure}

\textbf{A Study of the Class-wise Accuracy for Super-classes.} 
With the assumption that confusion is more likely to happen between classes with similar semantic features, such as different dog breeds, we perform a study on CIFAR100 consisting of 20 super-classes, each having 5 sub-classes with semantically similar features. Figure~\ref{fig:confusion} (left) shows that most confusions are located close to the diagonal, which already indicates that most confusions happen among the super-class. Note that the diagonal values were zeroed out, for better visualization. Further, since the CIFAR-100 dataset is not semantically sorted, we sort the class indices according to the CIFAR-100 super-classes. To further investigate this, we zoom in into the ``people" super-class, for which the confusion matrix is shown in Figure~\ref{fig:confusion} (center) as well as the confusion from the ``people" super-class to the ``reptiles" super-class in Figure~\ref{fig:confusion} (right). It can be observed that the confusions are significantly higher among classes belonging to the super-class than towards a different super-class, where nearly no confusions can be observed.

\begin{figure}[htpb!]
    \centering
    \includegraphics[width=0.45\linewidth]{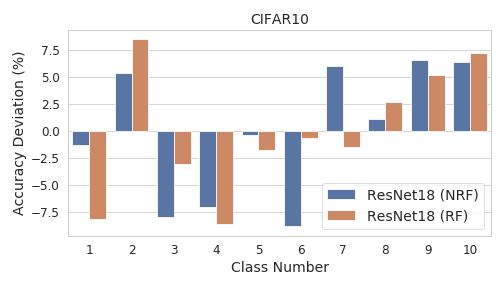}
    \includegraphics[width=0.35\linewidth]{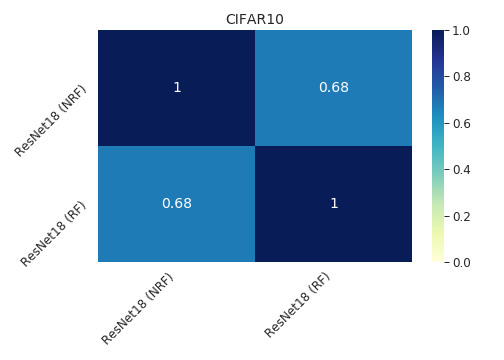}
    \caption{Class-wise accuracy deviation (left) and cosine similarity (right) for a ResNet18 trained on non-robust features (NRFs) and robust features (RFs).}
    \label{fig:nrf_rf}
\end{figure}

\textbf{Investigation from the Feature Perspective.} Inspired by~\citep{ilyas2019adversarial, benz2020data}, we generate datasets containing mainly non-robust features (NRFs) and robust features (RFs) from a standard model and an adversarially trained model, respectively. Figure~\ref{fig:nrf_rf} presents a ResNet18 trained NRFs and RFs~\citep{madry2017towards}. The results suggest that the class-wise discrepancy is observed on both, suggesting the robustness of the dataset features does not have a significant influence on the class-wise accuracy distribution. 

\subsection{Attempts to Mitigate the Class-wise Imbalance}

\textbf{Towards Fair Standard and Robust Training.} We attempted to use the values of the confusion matrix as a regularizer to decrease the overall class-wise accuracy imbalance. Unfortunately, this approach did not lead to promising results, either significantly decreasing the overall accuracy or having only a limited effect on the class-wise imbalance. Here, we propose a simple approach to achieve a more balanced class-wise accuracy distribution. The method can be briefly summarized as a class-wise weighting of the cross-entropy loss. Hereby, the weighting adapts according to the class-wise accuracy deviations from the mean with a moving average. Specifically, given a weighting vector $\gamma \in \mathds{R}^{C}$, whereby $C$ indicates the number of classes and $\gamma_c$ indicates the weighting specific for class $c$. In the beginning, $\gamma$ is initialized to be all ones. After each epoch, the class-wise accuracies $\tau \in \mathds{R}^{C}$ and the overall accuracy $\zeta \in \mathds{R}$ are calculated. Based on the class-wise accuracy deviations from the mean, \ie\ ($\tau - \zeta$), the weighting values are updated with a value of $\alpha$, which we set to $0.05$. In short, the weighting value for a class is increased, if the class-wise accuracy is below the mean and the weighting value is decreased, if the class-wise accuracy is above the mean. Mathematically, this can be expressed as:
\begin{equation}
    \gamma_c = \gamma_c + \begin{cases} 
        \alpha  & \text{if } (\tau_c - \zeta) \leq 0 ,\\ 
        -\alpha & \text{if } (\tau_c - \zeta) > 0.
        \end{cases} 
\end{equation}

\begin{figure}[htpb!]
    \centering
    \includegraphics[width=0.32\linewidth]{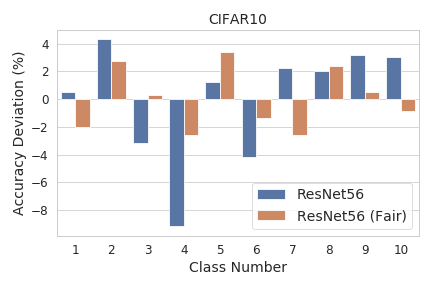}
    \includegraphics[width=0.32\linewidth]{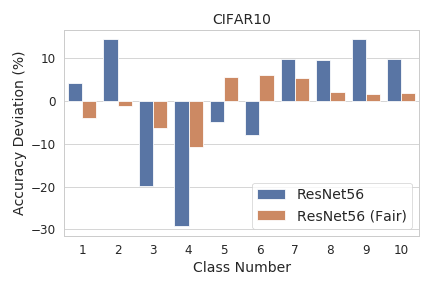}
    \includegraphics[width=0.32\linewidth]{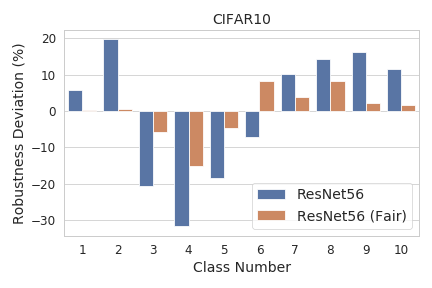}
    \caption{Comparison of the fair training method compared to normal model training for standard training (left) and adversarial training (center). For the adversarially trained models, the performance is also shown under adversarial attack (right).}
    \label{fig:fair}
\end{figure}

In Figure~\ref{fig:fair}, we present the results on ResNet56 for our fair weighting approach. For both standard training and adversarial training, the class-wise accuracy deviations are noticeably smaller than for the normal training without weighting. Specifically for normal standard training, the worst accuracy deviation is around $-9\%$ which is decreased to less than $-3\%$ for our fair weighting strategy. The improved fairness is at a small cost of overall accuracy, dropping from $92.9\%$ to $91.1\%$. Similar observations can be made for adversarial training where the worst accuracy deviation decreases from around $-30\%$ to around $-10\%$, at the cost of overall clean accuracy dropping from $78.9\%$ to $75\%$. Moreover, we are interested in whether the class-wise robustness under adversarial attack also gets more imbalanced, which is confirmed by the results in Figure~\ref{fig:fair} (right). For example, the worst robustness deviation decreases from more than $-30\%$ to around $-15\%$ with overall robust accuracy dropping from $48.8\%$ to $40.2\%$. It is worth mentioning that the robustness of most vulnerable class 4 increases by around $7\%$ with our fair training strategy.

\textbf{Increasing the Class-wise Accuracy of an Important Class.} In certain scenarios a specific class can be more important than others. For example in the case of autonomous driving the class ``pedestrian" might be more important to detect than other classes. Hence, for such a class a higher class-wise accuracy would be required. We test if the weighting strategy can also be used to increase only the accuracy of an important class. Therefore we do not update the weight, but keep them fixed, with pre-defined values. Previously, we observed that for CIFAR-10 the class with index $4$ usually exhibits the highest negative accuracy deviation from the mean accuracy. Hence, increasing class number $4$ is the most challenging scenario, which we adopt for the following experiments. We set all values of $\gamma$ to $1$, except for $\gamma_4$, which we set to a higher value. Table~\ref{tab:inc4} shows the influence of different $\gamma_4$ on the class-wise accuracy deviations and the cosine similarity. With increasing values of $\gamma_4$, an increase in the class-wise accuracy of $4$ can be observed. However, the average accuracies also decrease with increasing $\gamma_4$ values. 

\begin{table}[htpb!]
    \centering
    \small
    \begin{tabular}{c|ccccc}
        \toprule
        Class-wise weight for class $4$ ($\gamma_4$) & 1 & 10 & 20 & 50 & 100 \\
        \midrule
        Class-wise accuracy for class $4$ ($\tau_4$) & 83.7 & 84.1 & 84.6 & 93.1 & 97.0 \\
        Overall accuracy ($\zeta$)                   & 92.9 & 90.8 & 89.1 & 80.2 & 60.9 \\
        \bottomrule
    \end{tabular}
    \caption{Class-wise and overall accuracy result for increasing class with index $4$ for different values of $\gamma_4$.}
    \label{tab:inc4}
\end{table}

\begin{figure}[htpb!]
    \centering
    \includegraphics[width=0.6\linewidth]{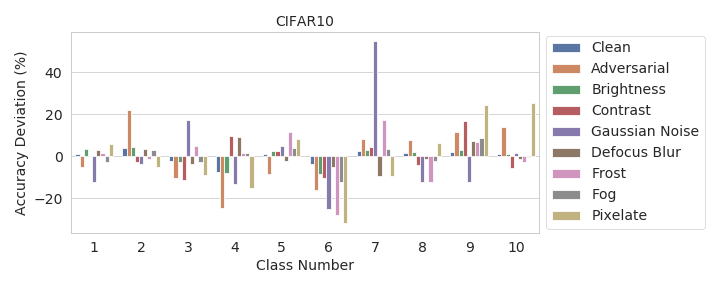}
    \includegraphics[width=0.39\linewidth]{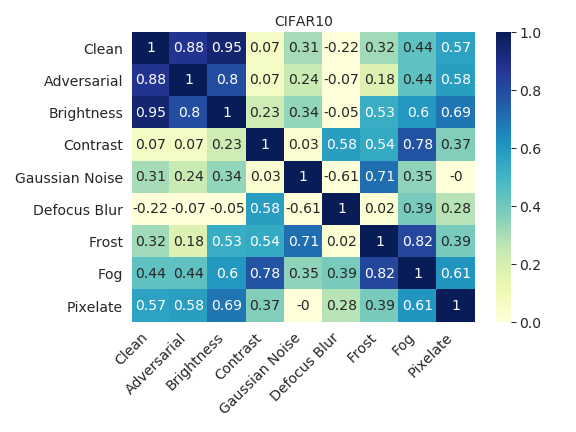}
    \caption{Class-wise accuracy deviation (left) and cosine similarity (right) for evaluating the influence of natural corruptions.}
    \label{fig:cifar10_corruption}
\end{figure}

\subsection{Additional Explorations for Robustness Against Natural Corruptions}
~\cite{hendrycks2019benchmarking} demonstrates that DNNs can be vulnerable to corruptions. The authors~\citep{hendrycks2019benchmarking} provided corrupted CIFAR10 datasets, namely CIFAR10-C, on which we evaluate the class-wise accuracy deviations. We present the results on CIFAR10 with ResNet56 for different corruptions in Figure~\ref{fig:cifar10_corruption}. Under corruptions, it is expected that the model accuracy significantly decreases. Specifically, the mean accuracies are $93.7\%$, $59.5\%$, $91.9\%$, $75.0\%$, $41.6\%$, $79.8\%$, $74.2\%$ and $86.5\%$ for without (clean), adversarial examples, brightness, contrast, Gaussian noise, defocus blur, frost, fog and pixelate corruption, respectively. It is worth mentioning that under adversarial perturbation, the most vulnerable class is still class 4, \ie\ cat, while it becomes class 6 for most natural corruptions. For example, with the natural frost and pixelate corruptions, class 6 is significantly more vulnerable than other classes. Another interesting phenomenon is that the accuracy of class 7 with Gaussian noise is strikingly higher than other classes and a significant amount of images from other classes are misclassified as class 7. 

\section{Conclusion}
Overall, this work performs an empirical study on the class-wise accuracy and robustness. We find that the phenomenon of inter-class discrepancy occurs in a wide range of setups, regardless of the dataset, model architecture, optimizer, learning schedule, etc. Adversarial training seems to make this inter-class discrepancy more serious, suggesting robustness might be at odds with inter-class fairness. The inter-class discrepancy can be, at least partially attributed to the fact that some classes share more similar features to other classes, causing confusion between those classes of similar features. Additionally, we find that the feature robustness, \ie\ robust features or non-robust features, does not have much to do with this discrepancy phenomenon. Finally, we attempt to mitigate this phenomenon by adaptively adjusting the class-wise loss weight. The side effect is that it decreases the overall accuracy or robustness, further suggesting inter-class fairness might be not easily achieved without the cost of overall performance. Increasing the accuracy of a certain important class also leads to an overall accuracy decrease.

\small
\bibliography{bib_mixed}

\end{document}